\documentclass[pdflatex,sn-apa,iicol]{sn-jnl}

\usepackage{graphicx}%
\usepackage{multirow}%
\usepackage{amsmath,amssymb,amsfonts}%
\usepackage{amsthm}%
\usepackage{mathrsfs}%
\usepackage[title]{appendix}%
\usepackage{xcolor}%
\usepackage{textcomp}%
\usepackage{manyfoot}%
\usepackage{booktabs}%
\usepackage{algorithm}%
\usepackage{algorithmicx}%
\usepackage{algpseudocode}%
\usepackage{listings}%
\usepackage{colortbl}
\theoremstyle{thmstyleone}%
\theoremstyle{thmstyletwo}%

\theoremstyle{thmstylethree}%

\begin{document}

\title[Article Title]{Rethinking Multi-Condition DiTs: Eliminating Redundant Attention via Position-Alignment and Keyword-Scoping}


\author[1]{\fnm{Chao} \sur{Zhou}}\email{chaozhou@mail.ustc.edu.cn}

\author*[2]{\fnm{Tianyi} \sur{Wei}}\email{tianyi.wei@ntu.edu.sg}

\author[1]{\fnm{Yiling} \sur{Chen}}\email{ustccyleric@mail.ustc.edu.cn}

\author[1]{\fnm{Wenbo} \sur{Zhou}}\email{welbeckz@ustc.edu.cn}

\author[1]{\fnm{Nenghai} \sur{Yu}}\email{ynh@ustc.edu.cn}

\affil[1]{\orgdiv{School of Cyber Science and Technology}, \orgname{University of Science and Technology of China}, \orgaddress{\street{Fuxing Road 100}, \city{Hefei}, \postcode{230026}, \state{Anhui}, \country{China}}}

\affil[2]{\orgdiv{College of Computing and Data Science}, \orgname{Nanyang Technological University}, \orgaddress{\street{50 Nanyang Avenue}, \city{Singapore}, \postcode{639798}, \state{Singapore}, \country{Singapore}}}


\abstract{
While modern text-to-image models excel at prompt-based generation, they often lack the fine-grained control necessary for specific user requirements like spatial layouts or subject appearances. 
Multi-condition control addresses this, yet its integration into Diffusion Transformers (DiTs) is bottlenecked by the conventional ``concatenate-and-attend'' strategy, which suffers from quadratic computational and memory overhead as the number of conditions scales.
Our analysis reveals that much of this cross-modal interaction is spatially or semantically redundant.
To this end, we propose Position-aligned and Keyword-scoped Attention (PKA), a highly efficient framework designed to eliminate these redundancies. 
Specifically, Position-Aligned Attention (PAA) linearizes spatial control by enforcing localized patch alignment, while Keyword-Scoped Attention (KSA) prunes irrelevant subject-driven interactions via semantic-aware masking. 
To facilitate efficient learning, we further introduce a Conditional Sensitivity-Aware Sampling (CSAS) strategy that reweights the training objective towards critical denoising phases, drastically accelerating convergence and enhancing conditional fidelity.
Empirically, PKA delivers a 10.0$\times$ inference speedup and a 5.1$\times$ VRAM saving, providing a scalable and resource-friendly solution for high-fidelity multi-conditioned generation.
}




\maketitle
\section{Introduction}

\begin{figure*}[tp]
    \centering
    \includegraphics[width=\linewidth]{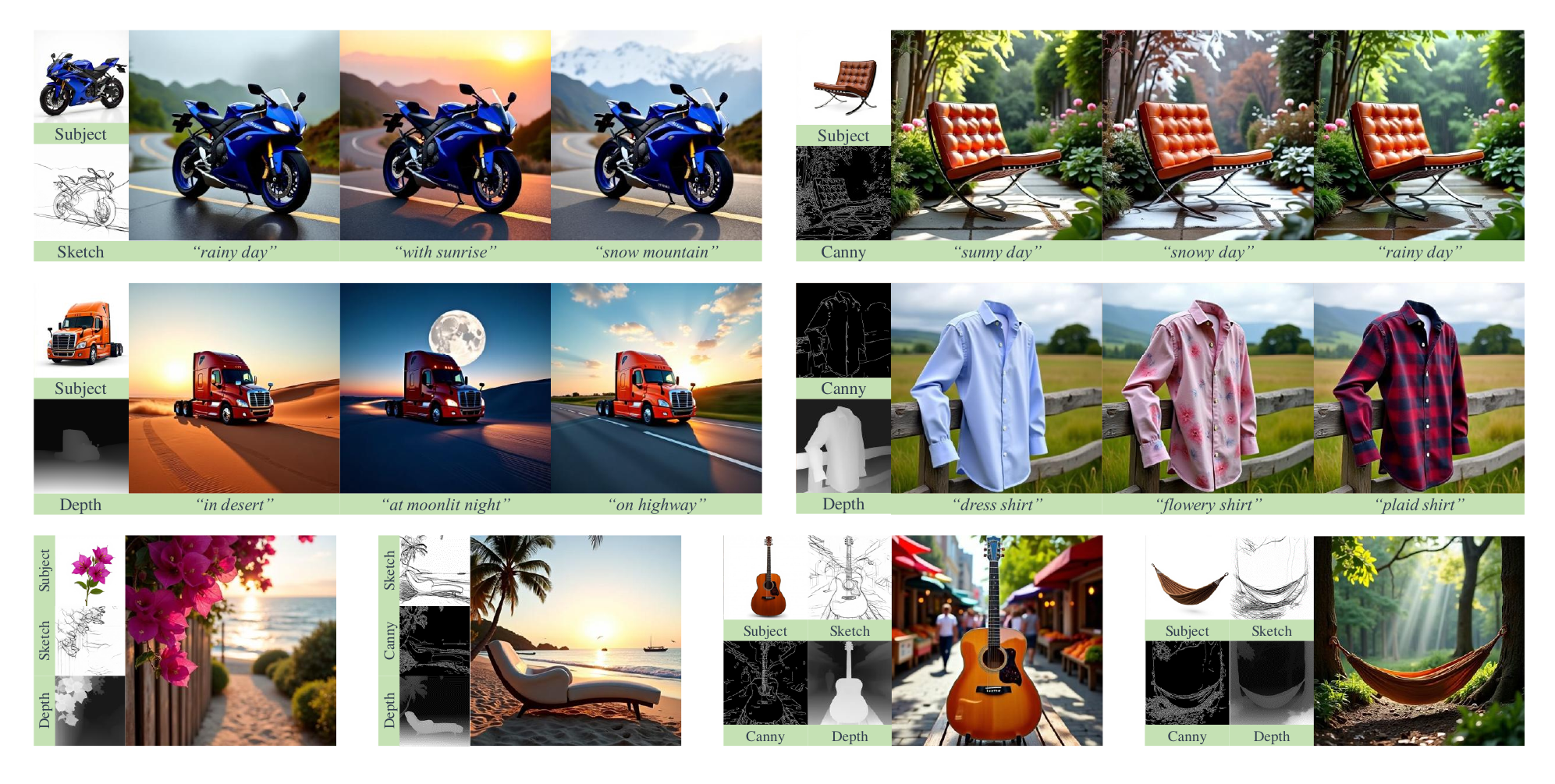}
    \caption{{Visual results of our proposed PKA on multi-conditional generation.} Our proposed PKA achieves high-quality multi-conditional generation with remarkable efficiency. Zoom in for better visualization.}
    \label{fig:visual_results}
\end{figure*}

After years of rapid development, Diffusion Transformers (DiTs)~\citep{Peebles2022DiT, esser2024scaling} have become a leading architecture for image generation.
While their performance is remarkable, most existing DiTs are guided predominantly by textual prompts. In many real-world scenarios, users often require more fine-grained control, such as specifying spatial arrangements, layouts, or visual references. This calls for multi-condition diffusion models that can flexibly incorporate both textual prompts and visual conditions.

In UNet-based diffusion models~\citep{podell2023sdxlimprovinglatentdiffusion}, this challenge is typically addressed via feature-level fusion, as exemplified by methods like ControlNet~\citep{zhang2023adding}, in which different condition modalities are injected at various layers of the UNet via feature addition or modulation~\citep{he2024dynamiccontrol, he2025anystoryunifiedsinglemultiple}. Since feature fusion is less straightforward in transformer architectures, DiTs typically adopt a different paradigm: an attention-based interaction where all condition and noisy image tokens are concatenated and processed jointly~\citep{yang2025image}. 

However, this ``concatenate-and-attend'' strategy is computationally prohibitive. Assuming $c$ condition inputs and $n$ tokens per condition, the resulting attention computation scales as 
$\mathcal{O}(c^2n^2)$ due to the pairwise attention across all conditions and noisy image tokens at each transformer block. As the number of conditions increases (e.g., layout, reference image, and depth maps), the total sequence length grows substantially. The attention mechanism's computational and memory demands scale quadratically, creating a critical bottleneck that leads to excessive memory consumption and inference latency.
This naturally forces a central question: \textit{Does effective multi-condition control truly require such massive attention computation?} 

\begin{figure}[h]
    \centering
    \begin{minipage}{0.48\textwidth}
        \centering
        \includegraphics[width=\linewidth]{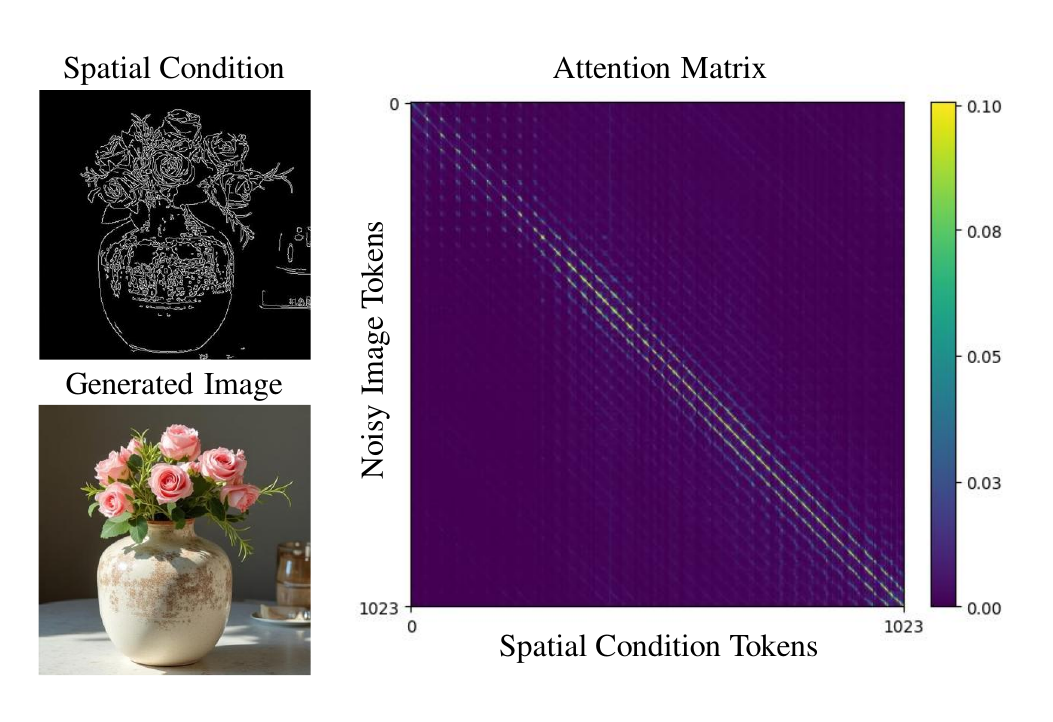}
        \caption{Visualization of the attention matrix in spatial-aligned generation. The heatmap exhibits a striking diagonal-dominant pattern, demonstrating that attention activations are primarily constrained to spatially congruent or proximal patches.}
        \label{fig:spatial2x}
    \end{minipage}
    \hfill % 添加水平间距
    \begin{minipage}{0.48\textwidth}
        \centering
        \includegraphics[width=\linewidth]{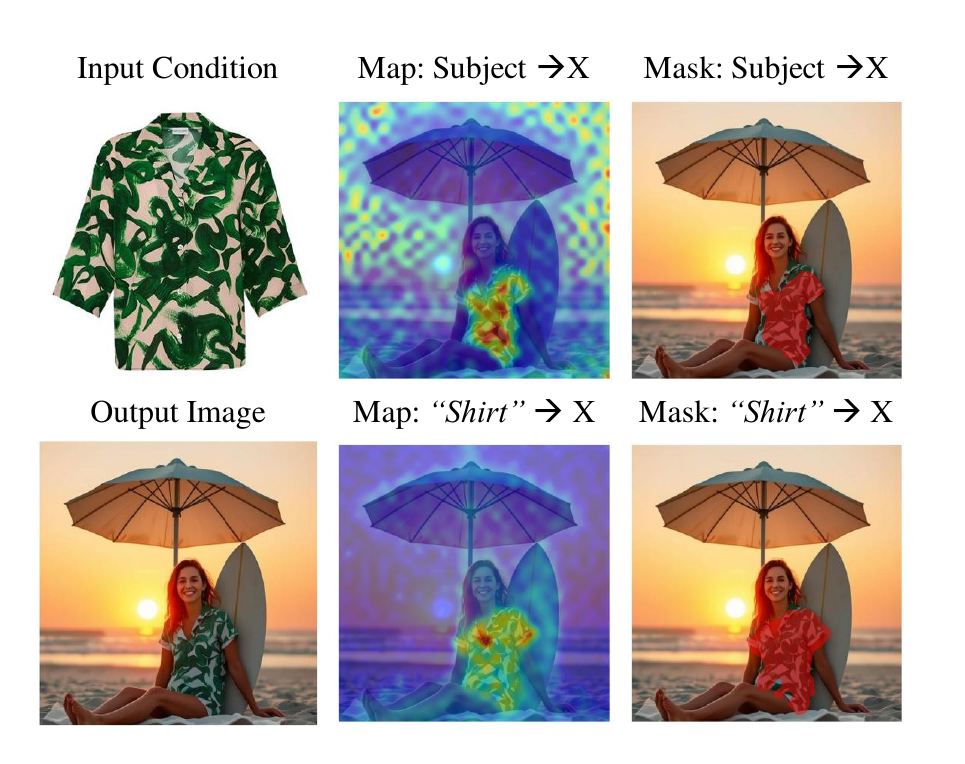}
        \caption{{Attention maps in subject-driven generation.} Prompt: ``On the beach, a lady wearing this shirt sits under a beach umbrella." X is the noisy image.}
        \label{fig:subject2x}
    \end{minipage}
    
\end{figure}

To answer this question, we systematically analyze the attention patterns within existing multi-condition DiTs~\citep{tan2024ominicontrol}. Our empirical investigation reveals that the ``concatenate-and-attend'' mechanism suffers from pronounced structural redundancy, which manifests in two distinct patterns.
For spatial-aligned conditions (e.g., layout maps), we observe that attention weights are strikingly concentrated along the diagonal of the attention matrix (Figure \ref{fig:spatial2x}). This indicates that meaningful cross-modal interactions are strictly confined to spatially congruent or proximal patches. The vast majority of off-diagonal entries—representing long-range, non-aligned interactions—contribute negligible activation energy, yet incur the same quadratic computational cost.
For subject-driven conditions (e.g., reference objects), the interaction exhibits a sparse activation profile. As illustrated in Figure \ref{fig:subject2x}, strong attention responses are limited to keyword-relevant regions, while the remaining image areas remain largely inactive.

Motivated by these observations, we propose Position-aligned and Keyword-scoped Attention (PKA), a novel mechanism for efficient multi-condition control. PKA leverages the inherent sparsity of attention patterns through two distinct, condition-aware components designed to eliminate computational waste.

For spatial-aligned conditions, Position-Aligned Attention (PAA) replaces full attention with a direct \textit{one-to-one} correspondence between noisy image and condition tokens at the same spatial coordinates. By computing attention only between these aligned pairs, PAA enables highly localized control with minimal overhead.

For subject-driven conditions, Keyword-Scoped Attention (KSA) is designed to prune irrelevant computations. It operates by first identifying the most relevant image regions via the naive detection ability of the attention map~\citep{zhou2025scale} between the textual keyword and the noisy image tokens. This map is then used to create a relevance-scoped mask, confining subsequent attention computations only to these \textit{salient regions} and drastically pruning the number of query-key interactions. Crucially, KSA leverages the emergent properties of the pre-trained attention maps and requires no additional design or modifications.

Furthermore, we contend that the standard log-uniform timestep sampling strategy, commonly adopted in Flow Matching~\citep{lipman2022flow} to bias training towards middle noise levels, remains insufficiently optimized for multi-conditional fine-tuning. While log-uniform sampling provides a generic density shift, it is agnostic to the specific temporal sensitivity of the model to heterogeneous visual conditions.

To refine this, we conducted a systematic perturbation analysis to characterize the conditional influence profile throughout the denoising trajectory. Our investigation reveals that the model's reliance on conditional signals is even more front-loaded than what log-uniform sampling accounts for, with sensitivity peaking sharply during the earliest structural-formation phases.
Consequently, we propose Conditional Sensitivity-Aware Sampling (CSAS), which strategically re-shapes the sampling distribution to further prioritize these high-sensitivity regimes. By aligning the training density with the intrinsic conditional dynamics revealed in our analysis, CSAS ensures more efficient gradient exploitation for complex control tasks, leading to faster convergence and superior conditional fidelity compared to standard log-uniform baselines.

By integrating these advancements, our experiments validate that we can significantly reduce both computational latency and the memory footprint of the attention mechanism, all without compromising the model's generative performance. Quantitatively, for scenarios with a high number of conditions, our method achieves an impressive speedup of up to 10.0$\times$ and a 5.1$\times$ reduction in memory consumption for the attention module. Some samples are shown in Figure \ref{fig:visual_results}.

In summary, our contributions are as follows.
\begin{itemize}
    \item We systematically characterize the spatial locality bias and semantic sparsity in multi-condition DiTs, providing an empirical foundation for eliminating redundancy inherent in the standard full-attention mechanism. 
    \item We propose the Position-aligned and Keyword-scoped Attention framework. By leveraging Position-Aligned Attention and Keyword-Scoped Attention, our method achieves nearly linear computational complexity.
    \item We introduce Conditional Sensitivity-Aware Sampling. By re-prioritizing the training objective toward high-sensitivity denoising phases, CSAS significantly accelerates fine-tuning convergence and enhances structural control fidelity.
    \item  We conduct comprehensive experiments, demonstrating that our method achieves state-of-the-art efficiency, including up to a 10$\times$ speedup while maintaining or even improving generation quality and controllability.
\end{itemize}

\section{Related Work}
\subsection{Controllable diffusion generation}
Multi-condition generation enables users to guide the synthesis process with diverse inputs like spatial layouts or reference subjects. In UNet-based architectures, this is often achieved via feature-level fusion. 
This versatile strategy has been successfully applied to a wide range of conditions. Prominent examples include ControlNet~\citep{zhang2023adding}, UniControl~\citep{qin2023unicontrol}, and UniControlnet~\citep{zhao2023uni}, which inject spatial guidance like edge maps or poses~\citep{zheng2024mmot, wang2025autostory, lin2024ctrlx, li2024controlnet++,zhang2025objctrl}. Following the same principle, methods like IP-Adapter~\citep{ye2023ipadapter}, Styleadapter~\citep{wang2025styleadapter}, EZIGen~\citep{duan2024ezigen}, and InstantID~\citep{wang2024instantid} utilize feature fusion to incorporate subject appearance from reference images, thereby ensuring identity consistency~\citep{jin2025unicanvas, li2025diffusion, dong2025dreamartist}. Despite their empirical success, these UNet-based approaches often rely on ad-hoc feature fusion strategies. This introduces architectural complexity and limits scalability, particularly when combining multiple, heterogeneous conditions.

In contrast, DiT-based models typically achieve multi-condition control through attention-based interaction. Frameworks like OminiControl~\citep{tan2024ominicontrol}  and UniCombine~\citep{wang2025unicombine} have demonstrated the viability of this paradigm, where all conditional and latent tokens are concatenated for joint processing through full self-attention~\citep{zhang2025creatilayout, zhang2025easycontrol}. However, this ``concatenate-and-attend'' approach faces a critical limitation: the computational cost grows quadratically with the number of tokens. This leads to substantial memory and runtime overhead, rendering these methods inefficient for practical scenarios that demand rich and varied conditional inputs.

\subsection{Efficient mechanism for diffusion transformers}
Several strategies have been proposed to mitigate the substantial computational overhead inherent in Diffusion Transformers (DiTs). One prominent research direction focuses on inference-time optimization, employing techniques such as temporal feature caching or the decomposition of redundant tokens to bypass repetitive computations~\citep{hunter2025fast, ma2024learningtocache, liu2025fastcachefastcachingdiffusion,chen2025accelerating}. Another orthogonal approach involves architectural streamlining, where efficiency is gained by identifying and removing or simplifying layers that contribute minimally to the terminal generation quality~\citep{fang2025tinyfusion, zhu2024dip, yang2025alter}. 
For specialized multi-condition generation, recent frameworks like PixelPonder~\citep{pan2025pixelponder} and OminiControl2~\citep{tan2025ominicontrol2} have introduced task-specific accelerations through dynamic token pruning and aggressive input down-sampling.

While effective, these approaches often compromise the fine-grained alignment between complex conditions and generated content. In stark contrast, our PKA module re-evaluates multi-conditional interactions from a first-principles perspective. Rather than relying on post-hoc token reuse or blind architectural pruning, we leverage domain-specific structural priors, namely the spatial congruence of layouts and the semantic locality of subjects, to eliminate redundancy at its source.

\section{Method}

\begin{figure*}[ht]
    \centering
    \includegraphics[width=\linewidth]{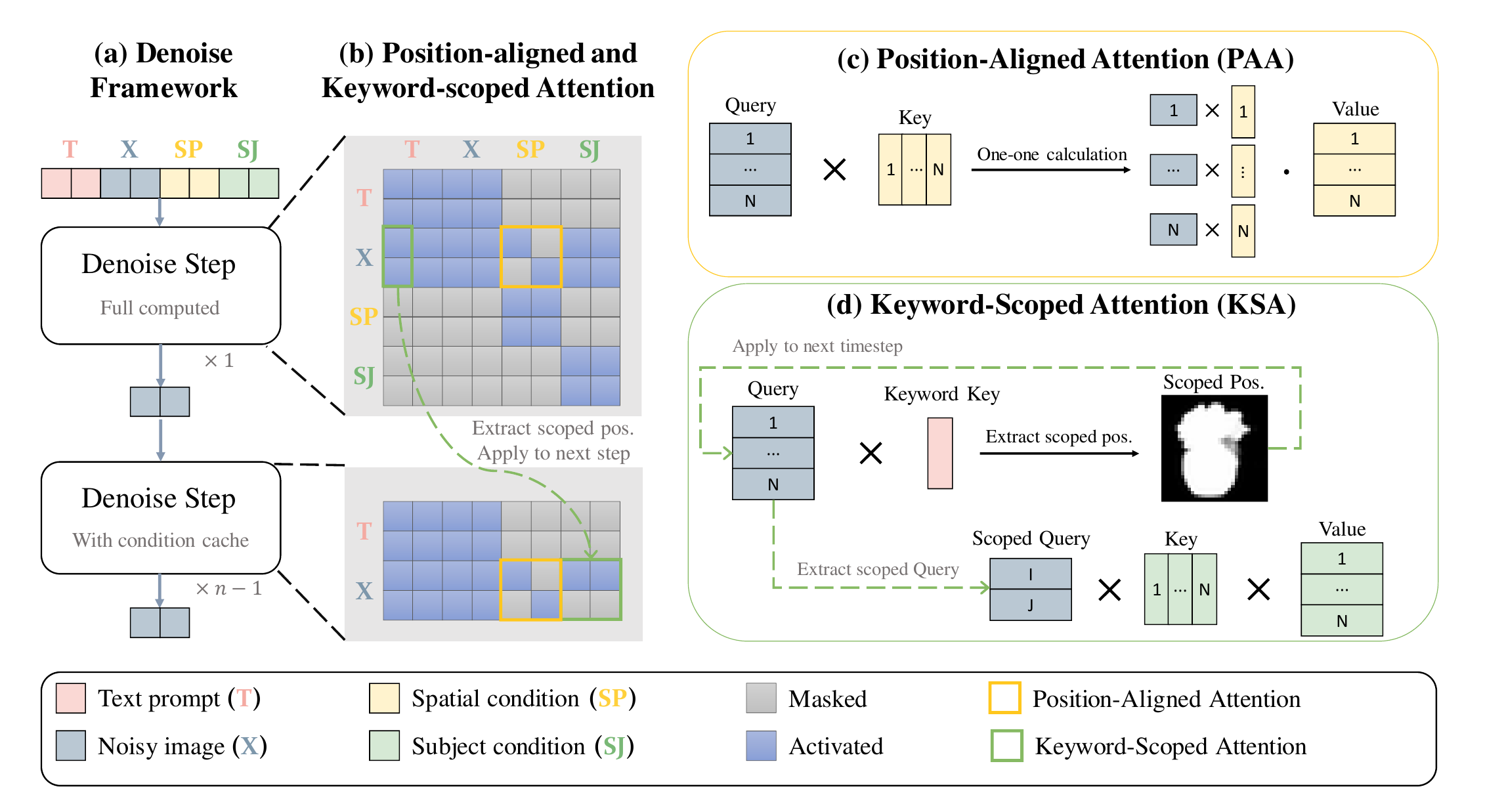}
    \caption{{Overview of our method.} 
    (a) The denoise framework. Full computation occurs only at the first step; the Keys and Values of all condition tokens are then cached for subsequent steps.
    (b) Position-aligned and Keyword-scoped Attention.Our decomposed attention mechanism, where conditions only perform self-attention (enabling the KV cache). The noisy image tokens (X) then interact with spatial (SP) and subject (SJ) conditions via PAA and KSA, respectively.
    (c) Position-Aligned Attention (PAA). PAA performs efficient one-to-one attention between the image (X) and spatial condition (SP) tokens at their aligned positions.
    (d) Keyword-Scoped Attention (KSA). KSA computes a relevance mask from text keywords in one step. This mask is then applied in subsequent steps to confine the attention computation between the image (X) and subject (SJ) to only the most relevant regions.
    }
    \label{fig:framework}
\end{figure*}

\subsection{Preliminary}

Diffusion Transformers (DiTs), such as FLUX.1~\citep{flux2024} and Stable Diffusion 3~\citep{esser2024scaling}, utilize a Transformer architecture as their denoising backbone. These models progressively refine noisy image tokens ($X \in \mathbb{R}^{N \times d}$), guided by various condition tokens like text ($C_T \in \mathbb{R}^{M \times d}$).

In multi-condition frameworks~\citep{tan2024ominicontrol}, additional visual condition tokens ($C_I \in \mathbb{R}^{N_I \times d}$) are incorporated by concatenating them with the text and image tokens. All tokens are then processed jointly through a multi-modal attention (MMA) mechanism:
\begin{equation}
\text{MMA}([C_T;X;C_I]) = \text{Softmax}\left(\frac{QK^\top}{\sqrt{d}}\right) V
\end{equation}
The primary issue with this ``concatenate-and-attend'' paradigm is its computational cost. The attention matrix $ QK^\top \in \mathbb{R}^{(M+N+N_I) \times (M+N+N_I)}$ scales quadratically with the sequence length, becoming prohibitively expensive as more conditions are added.

During training, these models typically use flow matching~\citep{lipman2022flow} learn the denoising process. Conventionally, the timestep $t$ for each training sample is drawn from a standard logit-normal distribution $\text{Logit-}\mathcal{N}(0, 1)$, ensuring the model is trained across all stages of the generation trajectory.

\subsection{Position-aligned and Keyword-scoped Attention}

\begin{figure}[ht]
    \centering
    \includegraphics[width=\linewidth]{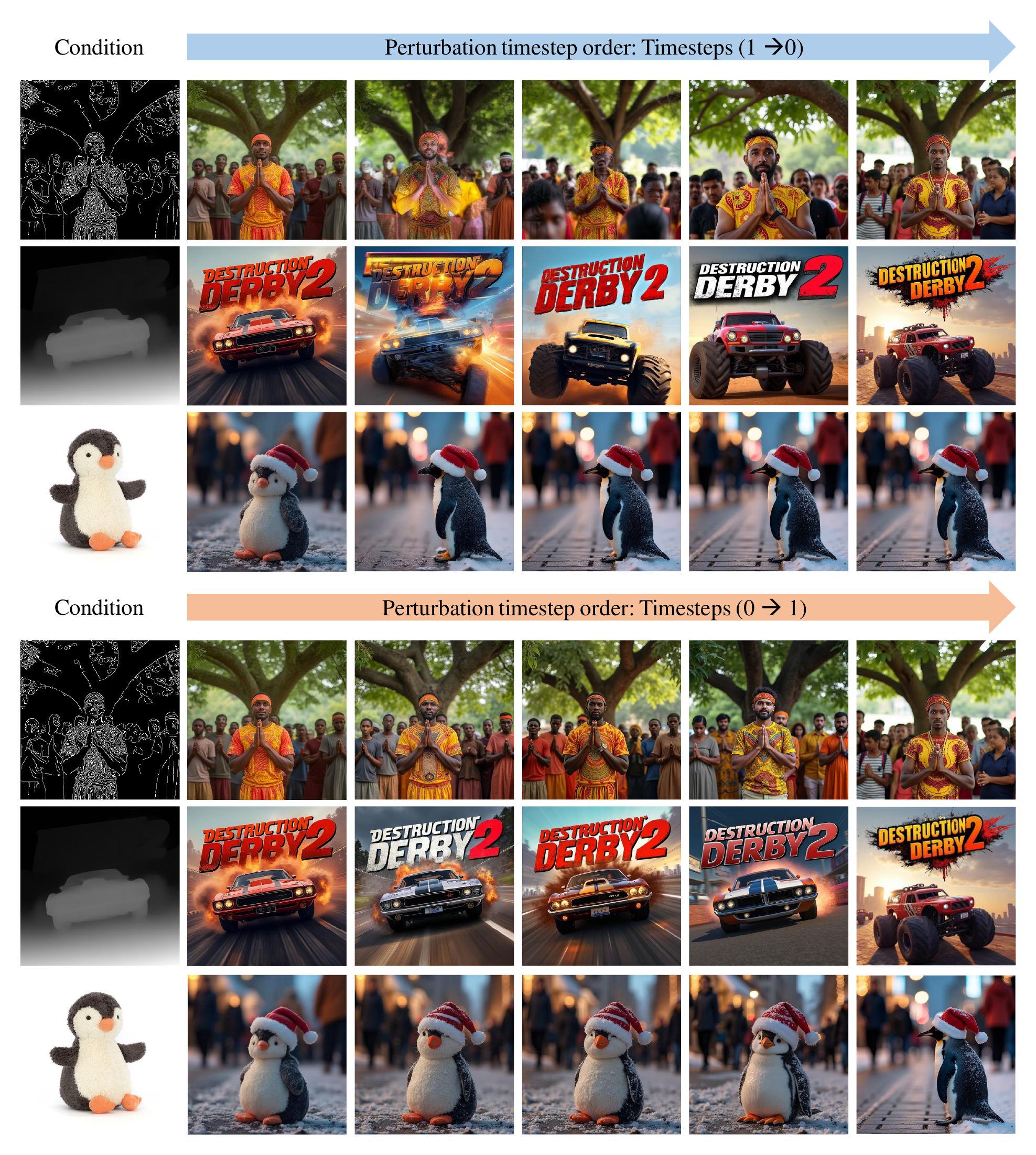}
    \caption{Qualitative results of visual condition perturbation. Left to right: visual condition, 0 (no perturbation), 7, 14, 21 perturbation steps, and 28 steps (no visual condition). Zoom in for better visualization.}
    \label{fig:perturbation_results_spatial}
\end{figure}

Building on the DiT-based text-to-image generation models, we propose  Position-aligned and Keyword-scoped Attention (PKA), a mechanism that decomposes the standard full-attention into a series of lightweight, specialized attentions. Our method operates on a sequence of tokens comprising text ($\mathbf{T}$), the noisy image ($\mathbf{X}$), the spatial condition ($\mathbf{SP}$), and the subject condition ($\mathbf{SJ}$). As illustrated in Figure \ref{fig:framework}(b), we fundamentally redesign the attention structure to reduce computational overhead. A key design principle is that condition tokens ($\mathbf{SP}$ and $\mathbf{SJ}$) only perform self-attention within their respective conditions. This structural choice enables a highly efficient condition cache mechanism, as shown in Figure \ref{fig:framework}(a). The Key and Value projections for all condition tokens are computed only once in the first denoising step and are then cached and reused for all subsequent steps. This eliminates redundant computations across the denoising trajectory. The noisy image tokens ($\mathbf{X}$) selectively interact with the conditions via our proposed Position-Aligned Attention (PAA) and Keyword-Scoped Attention (KSA) modules, while maintaining full attention with text ($\mathbf{T}$).

\subsubsection{Position-Aligned Attention}

For the spatial condition, we introduce Position-Aligned Attention (PAA). Our previous analysis reveals that the interaction for spatial control is diagonal-dominant, implying that long-range dependencies between non-aligned patches are redundant. Consequently, PAA reformulates the exhaustive global attention into a series of localized, position-wise interactions.

As illustrated in Figure \ref{fig:framework}(c), instead of attending to all condition tokens, each image token $\mathbf{X}_i$ at coordinate $i$ only interacts with its corresponding spatial condition token $\mathbf{SP}_i$. Formally, the PAA operation at position $i$ is defined as Equation \ref{eq:paa}.
\begin{equation}
\begin{aligned}
\text{PAA}(\mathbf{X}_i; \mathbf{SP}_{i}) &= \text{Softmax}\left(\frac{\mathbf{Q}_{X_i} \mathbf{K}_{SP_i}^\top}{\sqrt{d}}\right)\mathbf{V}_{SP_i}
\end{aligned}
\label{eq:paa}
\end{equation}
By enforcing this \textit{one-to-one} positional mapping, PAA effectively bypasses the $\mathcal{O}(N^2)$ complexity of the standard ``concatenate-and-attend'' mechanism. This design linearizes the computational overhead to $\mathcal{O}(N)$, enabling the model to incorporate an arbitrary number of spatial conditions without the prohibitive memory growth typical of standard DiTs.

\subsubsection{Keyword-Scoped Attention}
For the subject condition, we propose Keyword-Scoped Attention (KSA) to exploit the semantic sparsity. Our key insight is that the influence of a specific subject is typically confined to its localized semantic footprint rather than the entire canvas. Consequently, standard global attention incurs significant computational waste by attending to irrelevant background regions.

To address this, KSA introduces a dynamic masking strategy grounded in the temporal consistency of the denoising trajectory~\citep{zhou2025scale}. As illustrated in Figure \ref{fig:framework}(d), we employ a decoupled mask-generation mechanism. In timestep $t$, as Equation \ref{eq:mask}, we first derive a binary semantic mask $\mathbf{M}^t$ by capturing the affinity between image queries $\mathbf{Q}_X^t$ and a sparse set of keyword tokens $\mathbb{K}$, which anchor the subject's identity.
\begin{equation}
\mathbf{M}^t = \mathbb{I}\left[\text{Softmax}\left(\sum_{T_i\in \mathbb{K}}\frac{ \mathbf{Q}^t_X{\mathbf{K}_{T_i}^t}^\top}{\sqrt{d}}\right) \ge \epsilon\right]
\label{eq:mask}
\end{equation}
where $\mathbb{I}[\cdot]$ is the indicator function and $\epsilon$ denotes the activation threshold. In practice, $\mathbb{K}$ is extremely sparse (typically 1--2 tokens), ensuring negligible overhead for mask generation. Unless otherwise specified, we use $\epsilon = 0.2$ in the experiments.

Leveraging the observation that the semantic layout remains stable across adjacent timesteps, we apply $\mathbf{M}_t$ to the subsequent step $t+1$ to prune the attention manifold. The final KSA operation is only executed on the activated image tokens $\hat{\mathbf{Q}}_X^{t+1} = \mathbf{Q}_X^{t+1} \odot \mathbf{M}^t$, effectively restricting the subject's influence to its localized semantic footprint.
\begin{equation}
    \text{KSA}\left(\mathbf{X};\mathbf{SJ}\right) = \text{Softmax}\left(\frac{\hat{\mathbf{Q}}^{t+1}_X {\mathbf{K}_{SJ}^{t+1}}^\top}{\sqrt{d}}\right)\mathbf{V}^{t+1}_{SJ} 
    \label{eq:ksa}
\end{equation}

\subsection{Conditional Sensitivity-Aware Sampling}
Conventional Flow Matching frameworks~\citep{lipman2022flow} typically employ a logit-normal timestep sampling strategy, where $t \sim \text{Logit-}\mathcal{N}(0, 1)$, to bias the training toward mid-range noise levels. However, our investigation suggests that this generic distribution is suboptimal for fine-tuning multi-conditional control. As illustrated by the perturbation sensitivity analysis in Figure \ref{fig:perturbation_results_spatial}, the model~\citep{tan2024ominicontrol}'s reliance on conditional guidance is not uniform across the denoising trajectory; instead, it is heavily concentrated in the initial noise-dominant regime ($t \to 1$).

To align the optimization objective with this asymmetric sensitivity, we propose Conditional Sensitivity-Aware Sampling (CSAS). Rather than uniform or standard logit-normal sampling, CSAS strategically re-shapes the sampling density to prioritize the high-noise structural formation phases. Formally, we draw timesteps from a parameterized shifted logit-normal distribution:
\begin{equation}
t \sim \text{Logit-}\mathcal{N}(\mu, \sigma^2), \quad \text{with } \mu > 0, \sigma > 1
\label{eq:csas}
\end{equation}
By shifting the mean $\mu$ toward the beginning of the ODE trajectory, CSAS concentrates the gradient signals on the temporal segments where conditional alignment is most critical. This targeted learning prevents the dissipation of training capacity on late-stage pixel refinement, thereby significantly accelerating convergence and enhancing structural controllability.

\begin{figure*}[h]
    \centering
    \includegraphics[width=\linewidth]{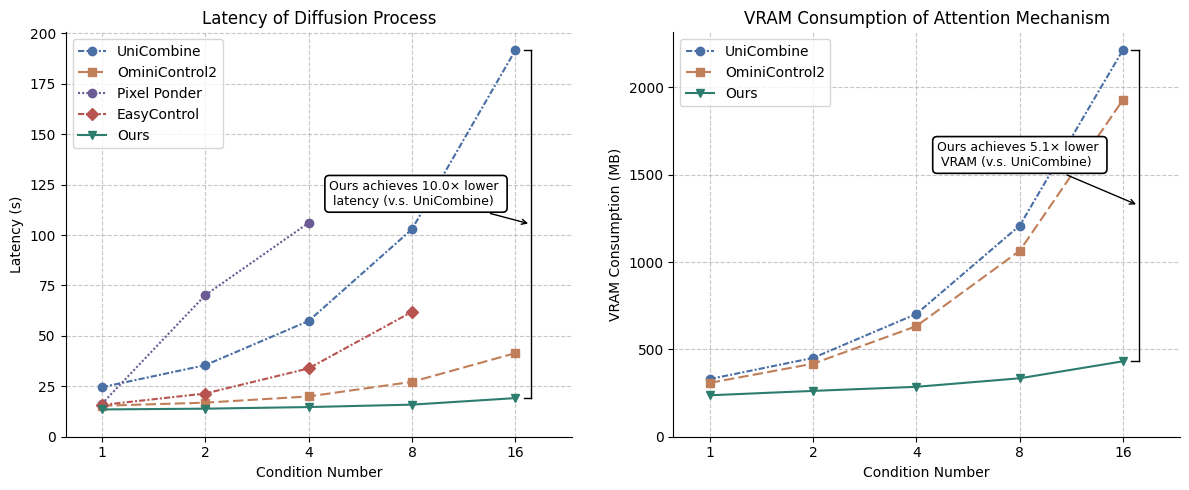}

    \caption{Scalability analysis of inference latency (left) and VRAM consumption (right) under varying spatial condition numbers.}
    \label{fig:latency_VRAM}

\end{figure*}

\section{Experiment}
\subsection{Setup}

\noindent\textbf{Training Details.}
% \noindent\textbf{Training Details.}
We curate a subset from the Subject200K dataset~\citep{tan2024ominicontrol}, ensuring each image caption contains a descriptive keyword. This subset is then partitioned into training and testing sets. To ensure a fair comparison, we fine-tune the FLUX.1~\citep{flux2024} model using LoRA~\citep{hu2022lora}, which is fine-tuned for 20,000 iterations using the Prodigy~\citep{mishchenko2023prodigy} optimizer with a batch size of 1 and a gradient accumulation step of 4.

\noindent\textbf{Evaluation Details.}
% \noindent\textbf{Evaluation Details.}
We align with previous work and conduct evaluations on subject-spatial condition control (Subject-Canny-to-Image and Subject-Depth-to-Image) and multi-spatial condition (Canny-Depth-to-Image). For the subject-spatial condition control task, we compare our method with OminiControl2~\citep{tan2025ominicontrol2}, UniCombine~\citep{wang2025unicombine}, and EasyControl~\citep{zhang2025easycontrol}. For the multi-spatial condition task, we additionally compare against PixelPonder~\citep{pan2025pixelponder} and UniControlnet~\citep{zhao2023uni}.
Efficiency metrics, including inference latency and attention VRAM consumption, are measured on a single NVIDIA RTX 6000 Ada GPU. 

\noindent\textbf{Metrics.}
To evaluate subject consistency, we calculate the CLIP-I~\citep{radford2021learning} and DINOv2~\citep{oquab2023dinov2} scores between generated images and ground-truth images. To measure controllability, we compute the F1 Score for edge conditions and the MSE score for depth conditions between maps extracted from the generated images and the original conditional inputs. For assessing generative quality, we compute FID~\citep{heusel2017gans} and SSIM~\citep{wang2004image} between the generated and ground-truth image sets. Additionally, we adopt the CLIP-T~\citep{radford2021learning} score to estimate the text consistency between the generated images and the text prompts.

\subsection{Main Results}

\subsubsection{Efficiency and Scalability}

To evaluate the computational efficiency and scalability of our framework, we conduct a controlled evaluation focusing on Position-Aligned Attention (PAA). We analyze the inference latency and attention-related VRAM consumption as the number of spatial conditions scales from 1 to 16, with results illustrated in Figure \ref{fig:latency_VRAM}. 
Note that we focus on PAA for this scalability test to ensure a deterministic comparison, as the overhead of Keyword-Scoped Attention (KSA) is inherently data-dependent (varying with subject size), a detailed discussion of which is provided in the Ablation Study.

Regarding inference latency (left plot), our method (green) maintains a nearly constant execution time, remaining almost unaffected by the increasing number of conditions. In stark contrast, existing methods such as UniCombine (blue), OminiControl2 (yellow), EasyControl (orange), and PixelPonder (purple) exhibit a significant, non-linear latency growth.
At the 4-condition mark, our method is approximately $4 \times$ faster than UniCombine, and $7 \times$ faster than PixelPonder. This efficiency gap widens to a factor of $10.0\times$ at 16 conditions, highlighting the superiority of our linearized attention design.
Notably, PixelPonder scales only up to 4 conditions, while EasyControl encounters out-of-memory (OOM) errors at 16 conditions.

In terms of VRAM consumption (right plot), we analyze the key bottleneck: the attention mechanism's overhead. Our method excels here as well, with its memory consumption scaling nearly linearly ($\mathcal{O}(N)$) with the number of conditions. Conversely, both UniCombine and OminiControl2 suffer from the quadratic growth ($\mathcal{O}(N^2)$)inherent in the ``concatenate-and-attend'' paradigm, leading to an explosive demand for memory. At 16 conditions, our method consumes $5.1\times$ less VRAM than UniCombine. It is worth noting that we omit PixelPonder from this comparison, as its adapter-based fusion design introduces prohibitive VRAM overhead, rendering it non-competitive in this multi-condition scenario.

\subsubsection{Qualitative comparison}

\begin{figure*}[tp]
    \centering
    \includegraphics[width=\linewidth]
    {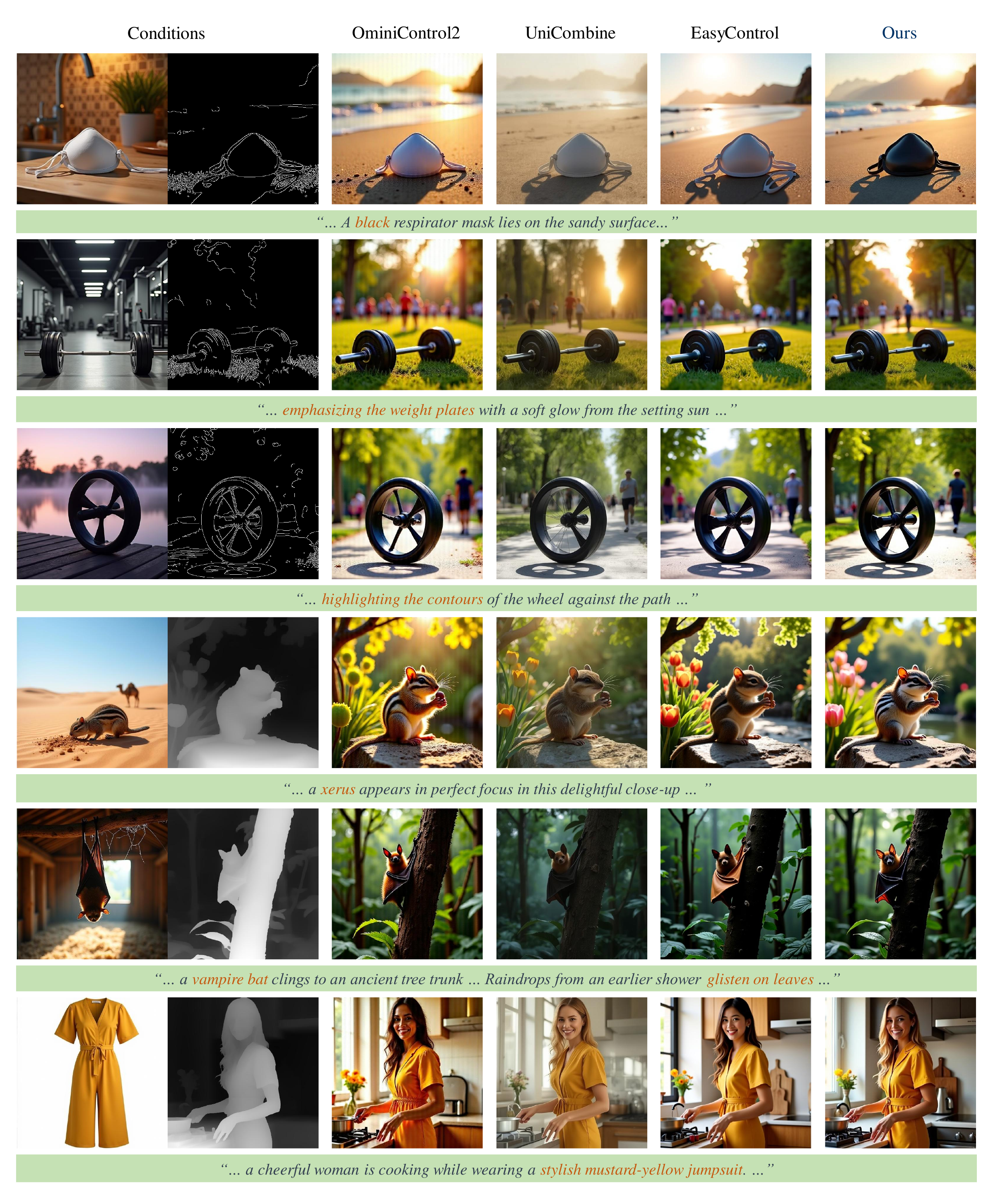}
    \caption{Qualitative comparison for subject-spatial conditions control. Our method achieves superior fidelity and higher image quality.}
    \label{fig:subject_spatial}
\end{figure*}

\begin{figure*}[tp]
    \centering
    \includegraphics[width=\linewidth]{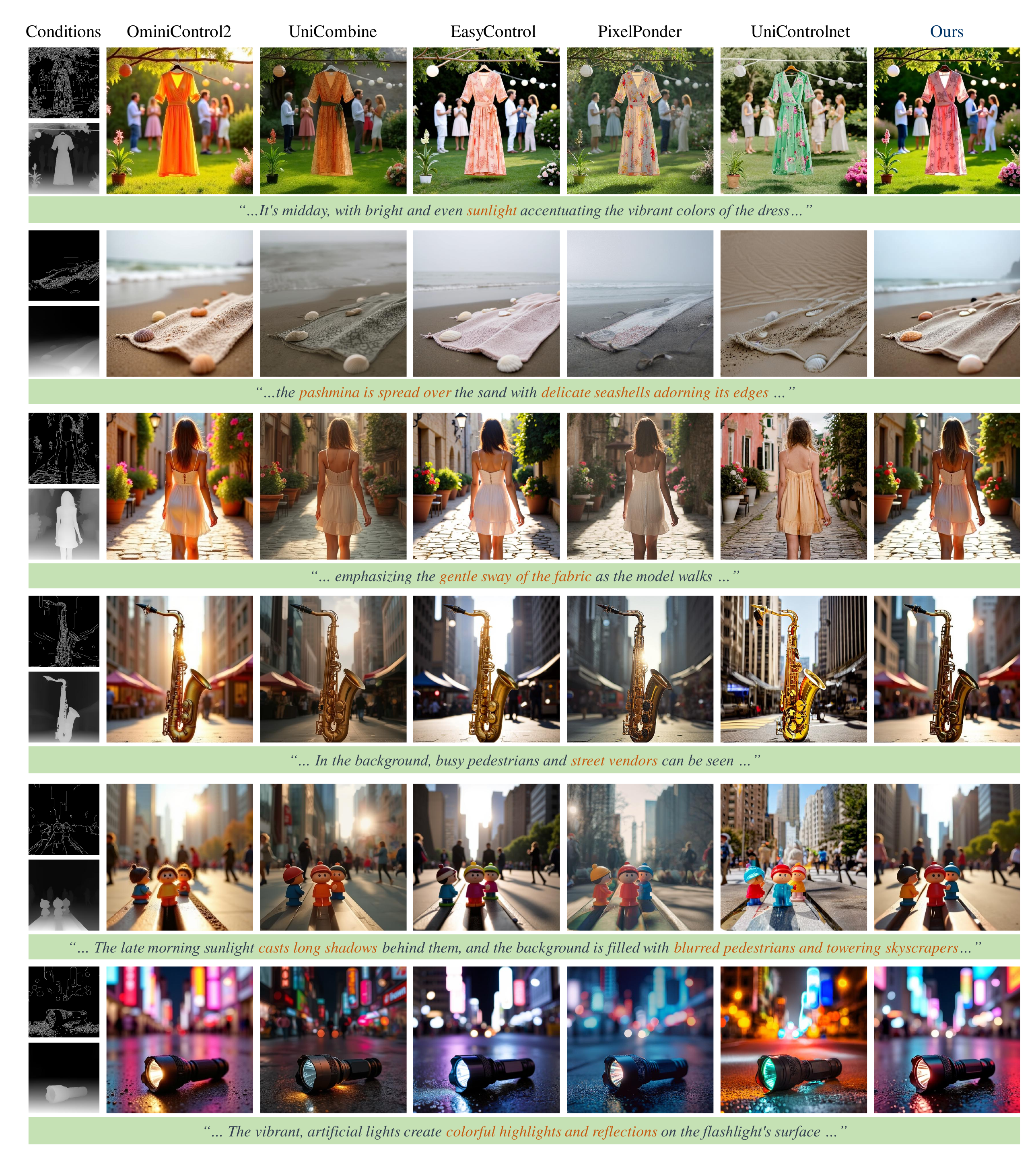}
    \caption{Qualitative comparison for multi-spatial conditions control. Our method demonstrates superior fidelity and higher image quality.}
    \label{fig:2spatial}
\end{figure*}

\begin{table*}[htp]
\centering
\small % 缩小表格内字体
\caption{Comparison of different methods across various tasks and metrics. The bold represents the optimal result.}
\label{tab:method_comparison}
\begin{tabular}{l|l|cc|cc|cc|c}
\toprule
\multirow{2}{*}{Task} & \multirow{2}{*}{Method} & \multicolumn{2}{c|}{Quality} & \multicolumn{2}{c|}{Controllability} & \multicolumn{2}{c|}{Consistency} & \multicolumn{1}{c}{Fidelity} \\
& & FID$\downarrow$ & SSIM$\uparrow$ & F1$\uparrow$ & MSE$\downarrow$ & CLIP-I$\uparrow$ & DINOv2$\uparrow$ & CLIP-T$\uparrow$ \\
\midrule
\multirow{4}{*}{Subject-Canny} & OminiControl2 & 72.03 & 0.406 & 0.192 & - & 0.878 & 0.867 & 0.327 \\
& UniCombine & 61.03 & 0.493 & \textbf{0.551} & - & 0.912 & 0.901 & \textbf{0.352} \\
& EasyControl & 57.53 & 0.506 & 0.287 & - & 0.935 & 0.915 & 0.347\\

&\cellcolor{cyan!10}Ours &\cellcolor{cyan!10}\textbf{52.99} &\cellcolor{cyan!10}\textbf{0.553} &\cellcolor{cyan!10}0.414 &\cellcolor{cyan!10}- &\cellcolor{cyan!10}\textbf{0.945} &\cellcolor{cyan!10}\textbf{0.926} &\cellcolor{cyan!10}0.349 \\
\midrule
\multirow{4}{*}{Subject-Depth} & OminiControl2 & 80.20 & 0.391 & - & 366 & 0.867 & 0.838 & 0.325 \\
& UniCombine & 70.22 & 0.454 & - & 312 & 0.911 & 0.879 & \textbf{0.350} \\
& EasyControl & 68.36 & 0.464 & - & 330 & 0.922 & 0.838 & 0.325\\
&\cellcolor{cyan!10}Ours &\cellcolor{cyan!10}\textbf{62.08} &\cellcolor{cyan!10}\textbf{0.515} &\cellcolor{cyan!10}- &\cellcolor{cyan!10}\textbf{160} &\cellcolor{cyan!10}\textbf{0.935} &\cellcolor{cyan!10}\textbf{0.904} &\cellcolor{cyan!10}0.348 \\
\midrule
\multirow{6}{*}{Multi-Spatial} & OminiControl2 & 71.87 & 0.475 & 0.194 & 303 & - & - & 0.342 \\
& UniCombine & 67.40 & 0.508 & 0.369 & 250 & - & - & \textbf{0.354} \\
& EasyControl & 62.38 & 0.516 & 0.291 & 233 & - & - & 0.344 \\

& PixelPonder & 65.60 & 0.506 & 0.385 & 321 & - & - & 0.350 \\
& UniControlnet & 90.71 & 0.337 & 0.269 & 563 & - & - & 0.345 \\

&\cellcolor{cyan!10}Ours &\cellcolor{cyan!10}\textbf{53.01} &\cellcolor{cyan!10}\textbf{0.613} &\cellcolor{cyan!10}\textbf{0.411} &\cellcolor{cyan!10}\textbf{114} &\cellcolor{cyan!10}- &\cellcolor{cyan!10}- &\cellcolor{cyan!10}0.353 \\
\bottomrule
\end{tabular}
\end{table*}

We present qualitative comparisons across two challenging scenarios: subject-spatial (Figure \ref{fig:subject_spatial}) and multi-spatial conditions (Figure \ref{fig:2spatial}) control. Comparing with others, our method achieves superior adherence to both the textual prompt and the visual conditions, consistently producing images with richer and more vibrant visual quality.

\begin{figure*}[h]
    \centering
    \begin{minipage}{\textwidth}
        \centering
        \includegraphics[width=\linewidth]{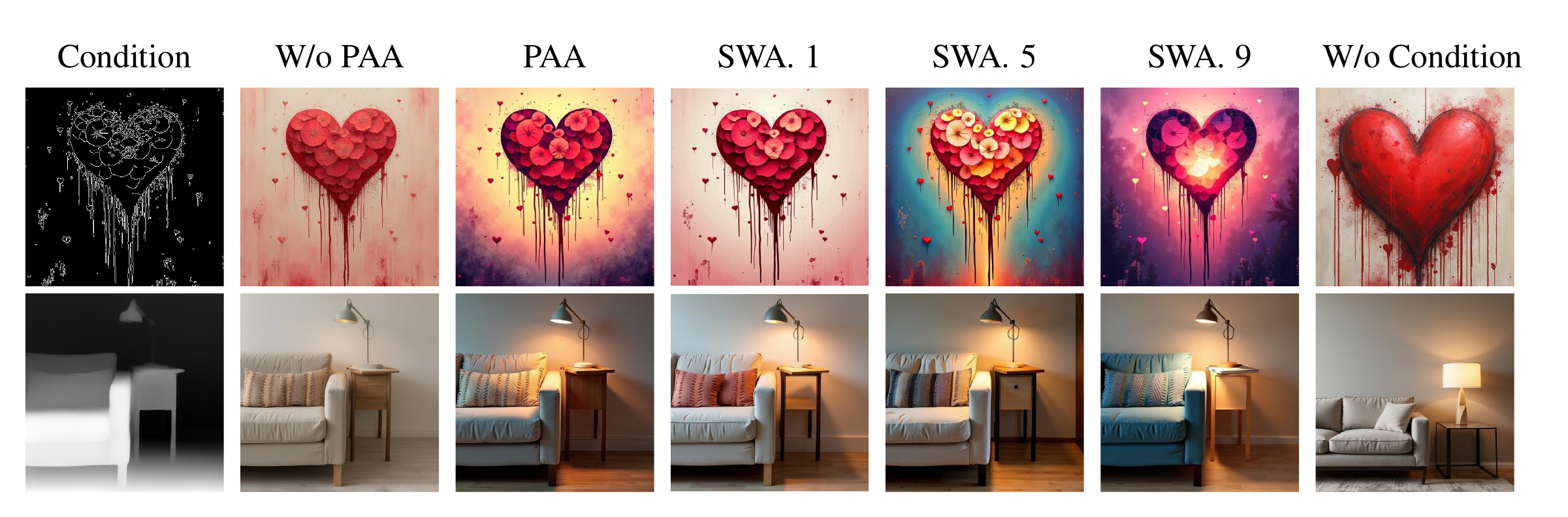}
        \caption{Qualitative ablation study of PAA. 
        We compare our Position-Aligned Attention (PAA) with Sliding Window Attention (SWA) of varying window sizes ($k \in \{1, 5, 9\}$).
        }
        \label{fig:PAA_ablation}
    \end{minipage}
    \hfill % 添加水平间距
    \begin{minipage}{\textwidth}
    \hfill
        \centering
        \vspace{1em}
        \includegraphics[width=\linewidth]{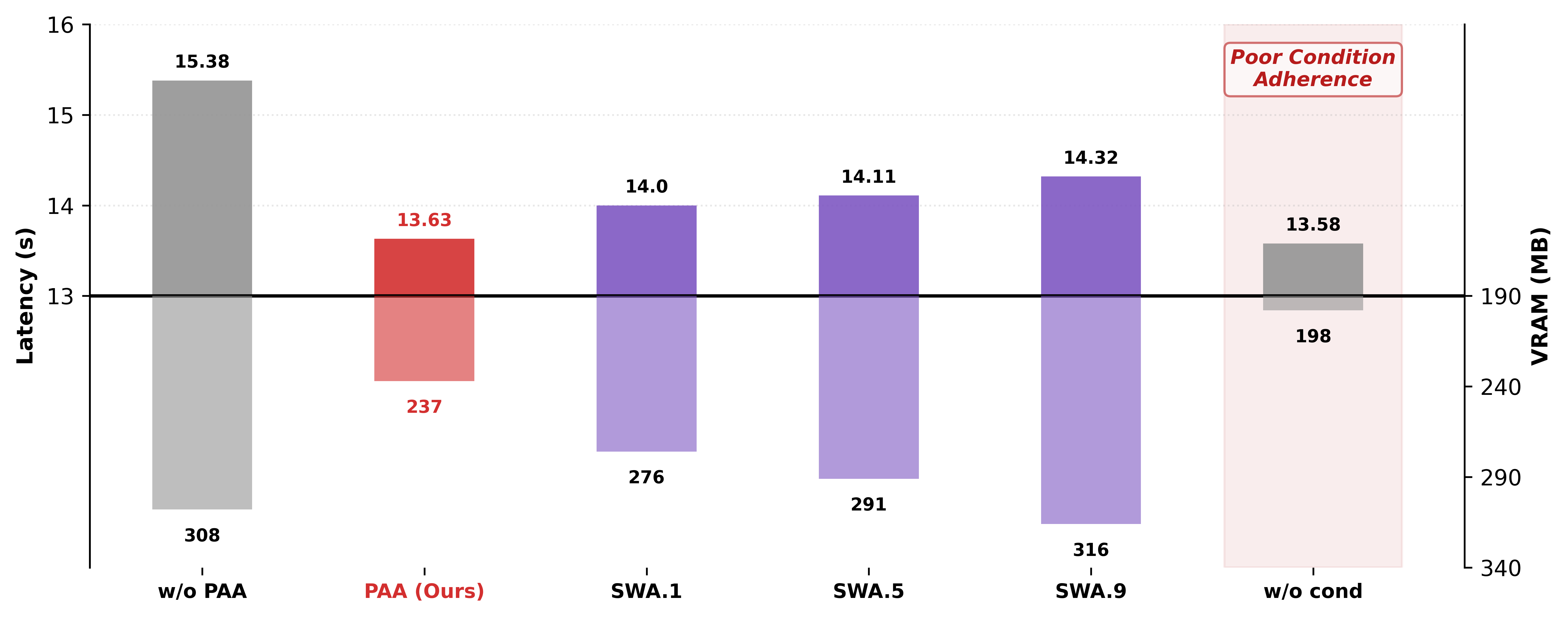}
        \caption{Efficiency-performance trade-off analysis of PAA. The bar chart illustrates the inference latency (top) and attention-related VRAM consumption (bottom) for different spatial interaction strategies.}
        \label{fig:Efficiency_PAA}
    \end{minipage}
\end{figure*}

As illustrated in Figure \ref{fig:subject_spatial}, our framework demonstrates a superior ability to simultaneously preserve the identity of reference subjects and adhere to spatial layouts. Compared to competing methods, which often suffer from semantic drifting or structural distortion under dual constraints, our model maintains high fidelity to both modalities. This is particularly evident in the third row of Figure \ref{fig:subject_spatial}, where our method is the only one to faithfully reconstruct the intricate Canny structures of \textit{the wheel and its corresponding shadow}, whereas others produce blurred or incomplete geometries. Similarly, in the fourth row, our model exhibits exceptional subject fidelity, precisely capturing the distinct high-frequency textures, such as \textit{the stripes of the xerus}, which are often lost or over-smoothed in the baseline results.

In scenarios involving multiple spatial conditions, existing models frequently encounter feature interference, leading to muted color palettes and diminished contrast. In contrast, our PKA framework avoids the noise injection typical of dense global attention by linearizing interactions. As shown in Figure \ref{fig:2spatial}, our model produces images with richer tonal range and vibrant color saturation compared to the baselines. The consistency between overlapping conditions is also significantly stronger, demonstrating that our lightweight design effectively balances multiple control signals without compromising the expressive capacity of the underlying DiT.

\subsubsection{Quantitative evaluation}

Table \ref{tab:method_comparison} presents a comprehensive quantitative evaluation across various multi-condition tasks. Our approach consistently outperforms state-of-the-art baselines in Generative Quality and Subject Consistency, while maintaining superior or highly competitive performance in Controllability and Text Fidelity.

Specifically, in terms of image quality, our method achieves the lowest FID scores across all categories, notably reaching 52.99 and 53.01 in Subject-Canny and Multi-Spatial tasks, respectively. This signifies a substantial improvement in visual realism over UniCombine and OminiControl2. Regarding Subject Consistency, our model leads the field with the highest CLIP-I and DINOv2 scores, demonstrating that the Keyword-Scoped Attention (KSA) effectively preserves the reference subject’s identity by focusing the denoising process on semantically relevant regions.

In terms of Controllability, our method establishes a new state-of-the-art in Subject-Depth and Multi-Spatial scenarios, with the MSE for depth control (114) being significantly lower than that of UniCombine (250). While UniCombine retains a marginal lead in F1 score for the Subject-Canny task, our method provides a superior balance between structural adherence and visual aesthetics, avoiding the over-saturation or artifacts often associated with aggressive control. Furthermore, our model preserves Text Fidelity (CLIP-T) at a level comparable to the leading baselines, with the slight numerical difference being perceptually indistinguishable. 

\subsection{Ablation Study}
\subsubsection{Effect of Position-Aligned Attention}
\begin{figure*}[h]
    \centering
    \begin{minipage}{\textwidth}
        \centering
        \includegraphics[width=\linewidth]{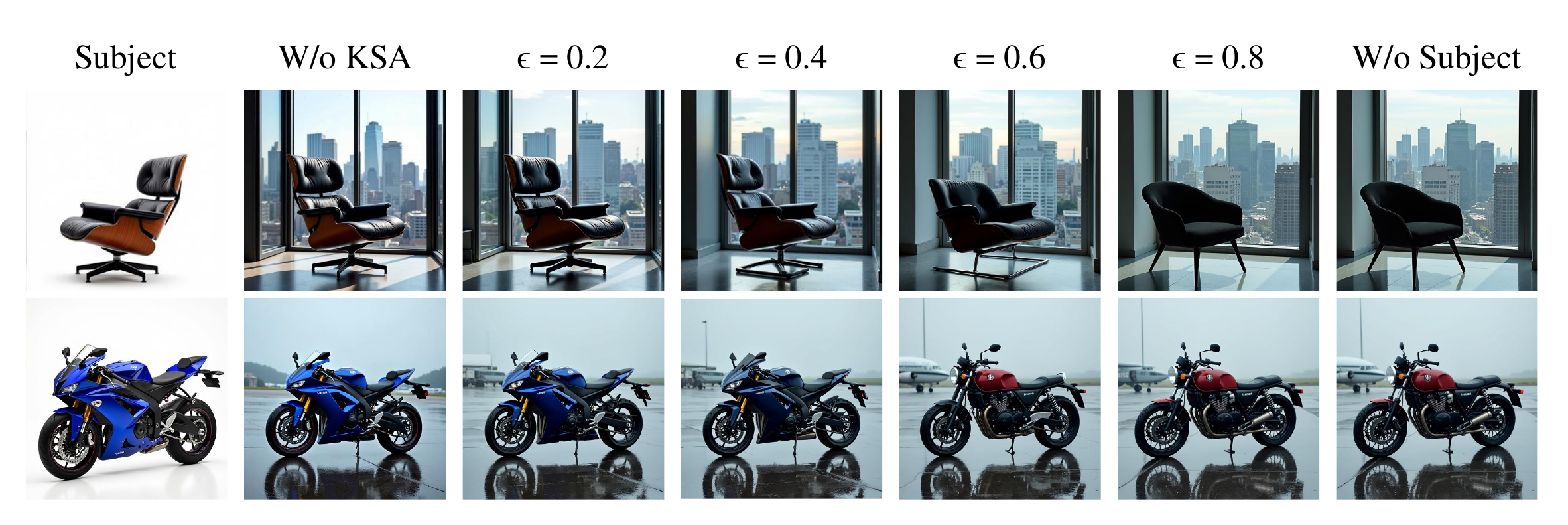}
        \caption{Visual ablation of the KSA mask threshold $\epsilon$. We evaluate the impact of the activation threshold on subject fidelity. As $\epsilon$ increases, the semantic mask becomes more restrictive, progressively filtering out peripheral subject features.}
        \label{fig:mask_ablation}
    \end{minipage}
    
    \hfill % 添加水平间距
    \begin{minipage}{\textwidth}
        \centering
        \vspace{1em}
        \includegraphics[width=\linewidth]{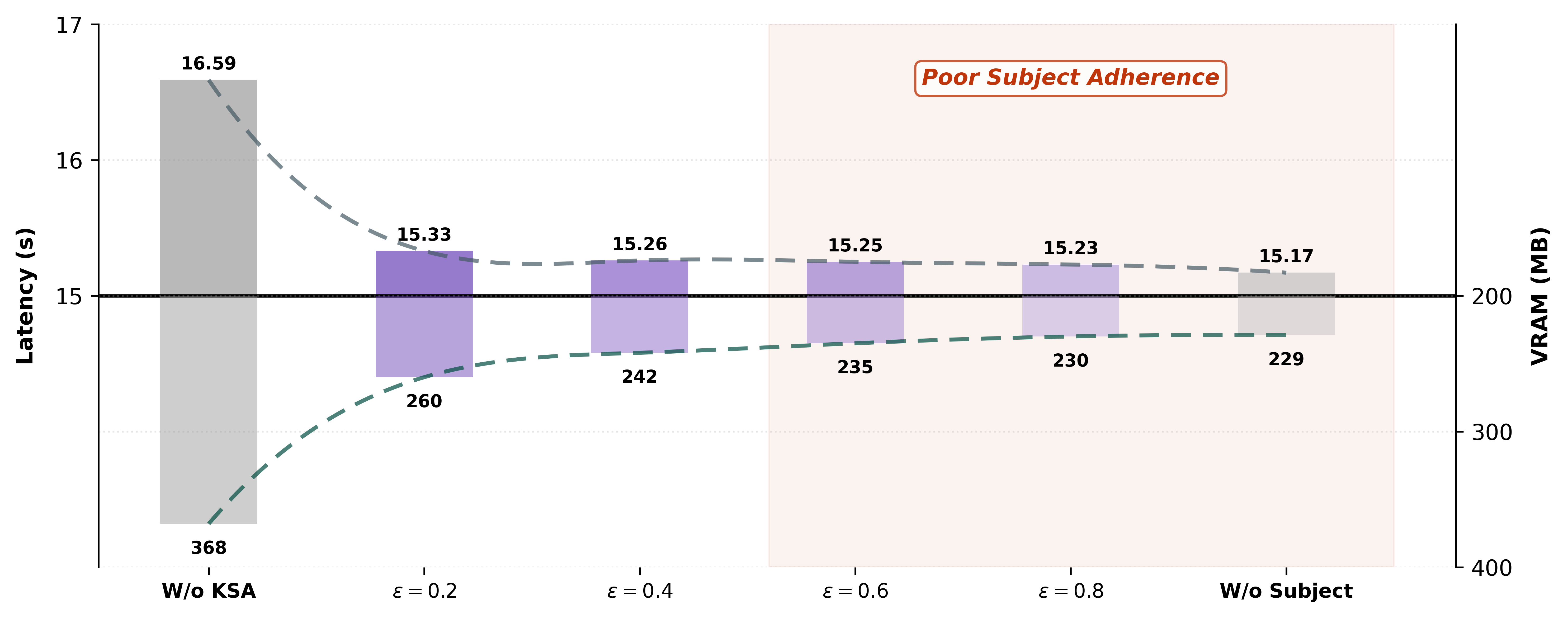}
        \caption{Efficiency and consumption scalability under varying KSA thresholds. The mirrored bar chart illustrates the trade-off between inference latency (top) and VRAM consumption (bottom).}
        \label{fig:Efficiency_KSA}
    \end{minipage}
\end{figure*}
To evaluate the efficacy and necessity of Position-Aligned Attention (PAA), we conduct a comparative study against two categories of baselines: Standard DIT (denoted as ``W/o PAA'') and Sliding Window Attention (SWA)~\citep{pan2023slide} with varying receptive fields ($k \in \{1, 5, 9\}$).

As qualitatively demonstrated in Figure \ref{fig:PAA_ablation}, all variants—including the standard full attention and the multi-scale SWA—produce high-fidelity images that strictly adhere to the provided spatial conditions (e.g., heart-shaped layout and interior geometry). This visual consistency across different attention scopes confirms a crucial empirical insight: the off-diagonal interactions in standard cross-attention are largely redundant for spatial grounding. The quantitative advantages of our design are further underscored in Figure \ref{fig:Efficiency_PAA}. While achieving comparable visual quality to the more complex SWA baselines, our PAA achieves the lowest inference latency (13.63s) and the most compact VRAM consumption (237MB).

\subsubsection{Effect of Keyword-Scoped Attention}

The Keyword-Scoped Attention (KSA) module introduces a principled mechanism to navigate the trade-off between computational efficiency and subject fidelity. By modulating the mask threshold $\epsilon$, KSA enables a tunable sparsity prior that selectively prunes cross-modal interactions.

\begin{figure*}[h]
    \centering
    \begin{minipage}{\textwidth}
        \centering
        \includegraphics[width=\linewidth]{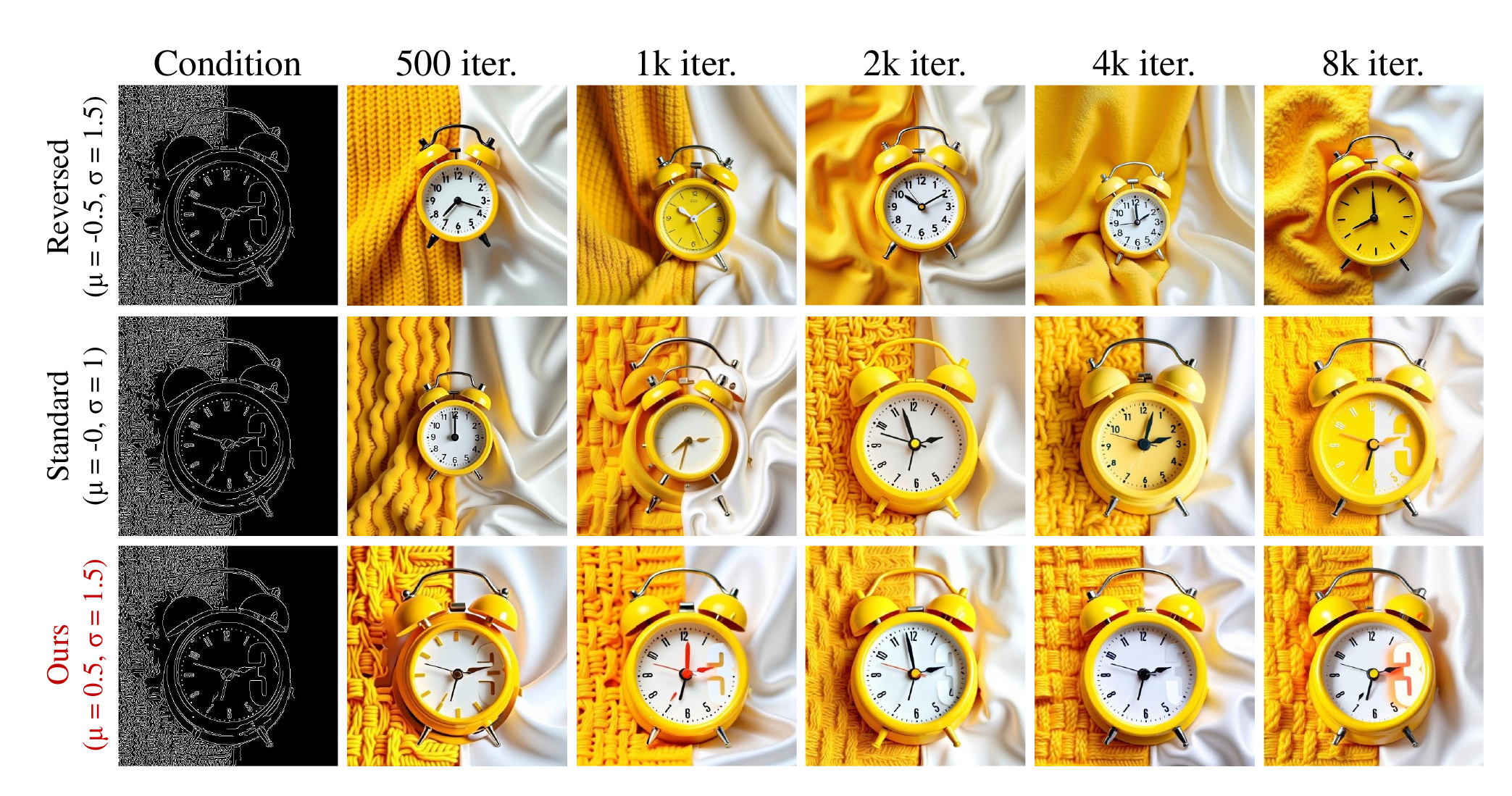}
        \caption{Qualitative ablation of the CSAS sampling strategy across fine-tuning iterations. We evaluate the structural grounding capability under different timestep sampling distributions ($\mu$ and $\sigma$).}
        \label{fig:early_step}
    \end{minipage}
    \hfill % 添加水平间距
    \begin{minipage}{\textwidth}
        \centering
        \vspace{1em}
        \includegraphics[width=\linewidth]{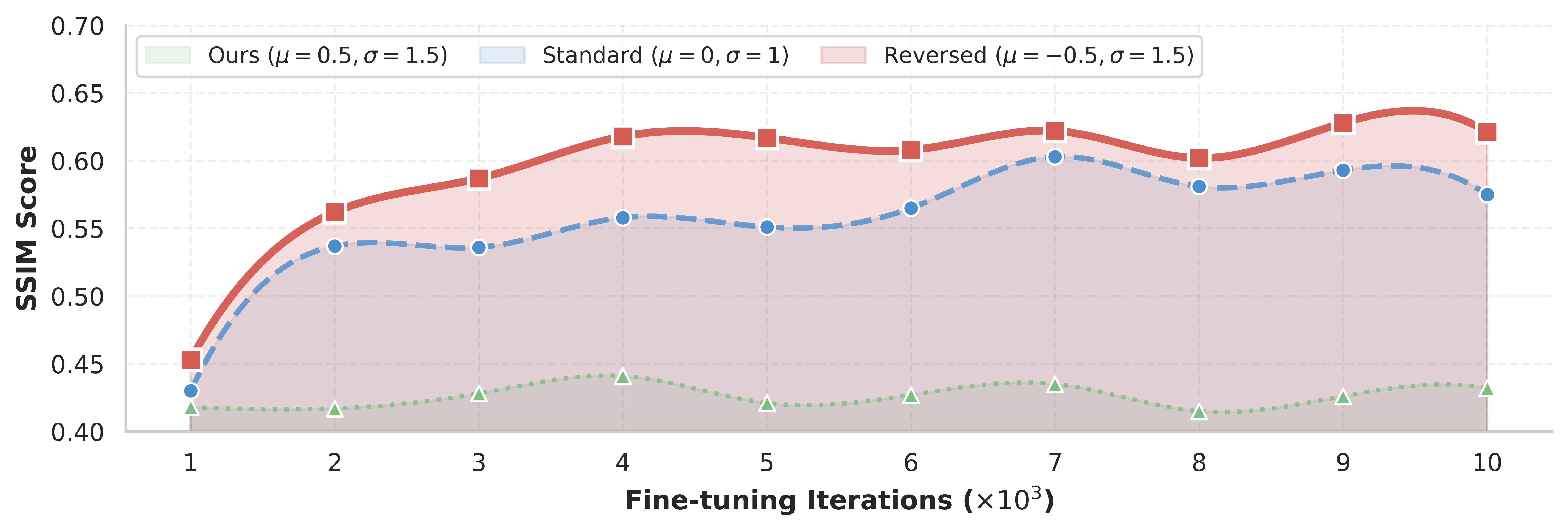}
        \caption{Quantitative convergence analysis of CSAS via SSIM scores. The plot illustrates the evolution of structural adherence over $10^3$ fine-tuning iterations.}
        \label{fig:early_step_line}
    \end{minipage}
\end{figure*}

As analyzed in Figure \ref{fig:Efficiency_KSA}, the baseline configuration (``W/o KSA'', equivalent to $\epsilon=0$) incurs a heavy computational burden, requiring 16.59s in latency and 368MB of VRAM. Upon activating KSA with $\epsilon=0.4$, we observe a significant Pareto improvement: VRAM consumption drops by 34.2\% (to 242MB), and latency is reduced to 15.26s. Remarkably, this substantial gain in efficiency does not lead to a catastrophic collapse of the reference identity. As qualitatively illustrated in Figure \ref{fig:mask_ablation}, the generated images remain highly congruent with the reference subjects even at aggressive threshold settings. The transition from $\epsilon=0$ to $\epsilon=0.4$ manifests as a graceful degradation of non-essential details, such as the precise curvature of the chair's legs or the subtle specular reflections on the motorcycle’s windshield, rather than a loss of core semantic features.

These results highlight the hyperparameter robustness of KSA, that $\epsilon$ can act as an intuitive control knob, allowing users to strategically allocate computational resources. There is a wide ``fidelity-stable'' region (where $\epsilon \in [0.2, 0.4]$) that KSA successfully identifies and eliminates semantic redundancies within the attention manifold.

\subsubsection{Effect of Conditional Sensitivity-Aware Sampling}

To evaluate the decisive impact of the sampling distribution on fine-tuning effectiveness, we compare three representative configurations for our CSAS strategy: Ours ($\mu=0.5, \sigma=1.5$, biased towards early high-noise stages), Standard ($\mu=0, \sigma=1$, standard logit-normal distribution), and Reversed ($\mu=-0.5, \sigma=1.5$, biased towards late low-noise stages).

The results in Figure \ref{fig:early_step} and Figure \ref{fig:early_step_line} provide overwhelming evidence of our method's superiority in both convergence speed and final generation quality. As illustrated by the convergence curves in Figure \ref{fig:early_step_line}, the slope of ours (red curve) significantly outperforms the baselines. ours achieves an SSIM score of $0.56$ at only $2\text{k}$ iterations, a performance level that requires more than double the training time (about $6\text{k}$ iterations) for the standard distribution. This ``fast-start'' advantage is qualitatively confirmed in Figure \ref{fig:early_step}, where ours is the only setting capable of reconstructing the alarm clock's geometric silhouette at the extremely early stage of $1\text{k}$ iterations. Beyond mere speed, ours attains a substantially higher saturation performance, with a final SSIM of $0.62$, markedly outperforming the standard sampling's $0.58$. In contrast, the Reversed strategy suffers from catastrophic learning failure, with SSIM scores stagnating below $0.45$.

\section{Conclusion}
In this paper, we addressed the computational inefficiency of multi-condition Diffusion Transformers by proposing Position-aligned and Keyword-scoped Attention (PKA), a novel mechanism that decomposes full attention into two efficient modules: Position-Aligned Attention (PAA) for spatial conditions and Keyword-Scoped Attention (KSA) for subject-driven ones. Our extensive experiments validate this approach, demonstrating a significant up to 10.0$\times$ inference speedup and a 5.1$\times$ reduction in VRAM consumption for the attention module, all while maintaining or even enhancing generative quality and controllability compared to state-of-the-art methods. Looking ahead, the significant efficiency gains of multi-condition control of PKA make it a promising foundation for tackling more complex generative tasks. A particularly exciting future direction is extending our framework to video generation, where PKA's principles could be applied to enforce temporal consistency across frames at a manageable computational cost. Ultimately, PKA offers a scalable and practical solution that paves the way for the next generation of complex and resource-friendly AI applications.
\section*{Statements and Declarations}

\subsection*{Funding}
This work was supported in part by the Natural Science Foundation of China under Grants 62372423, 62121002, 62072421, and was also supported by the Fundamental Research Funds for the Central Universities WK2100250070.

\subsection*{Conflict of Interest}
The authors declare that they have no competing interests.

\subsection*{Data Availability}
The dataset used in this paper was collected from publicly available online sources. We will release this dataset as open source upon acceptance of the paper to further benefit the research community. 

\subsection*{Code Availability}
The code used in this study will be released upon publication to support reproducibility and further research.

\subsection*{Author Contributions}
Chao Zhou and Tianyi Wei conceived the research idea, developed the proposed methodology, performed the experiments, and contributed to writing the manuscript. 
Yiling Chen, Wenbo Zhou, and Nenghai Yu provided scientific guidance, supervised the research design and experimental validation, and assisted in revising the manuscript. All authors reviewed and approved the final version of the manuscript.

\subsection*{Reproducibility}
We ensure reproducibility by providing clear distinctions between results and interpretations, along with detailed reporting of implementation components such as evaluation metrics, and computing resources. 

\subsection*{Ethical Considerations in Multi-condition Image Generation}
The advancement of multi-condition diffusion models, while offering significant creative potential, necessitates a rigorous discussion regarding ethical implications. Our proposed PKA framework, designed for lightweight subject-driven generation and spatial control, could theoretically be misused for the unauthorized creation of Synthetic Media or Deepfakes. By enabling precise control over character identity and spatial layouts, there is a risk that this technology could be employed to generate misleading content or infringe upon the Biometric Privacy and Intellectual Property of individuals without consent.

To mitigate these risks, we emphasize that our model is intended for academic research and creative assistance within ethical boundaries. We advocate for the integration of Digital Watermarking and Proactive Content Moderation protocols to ensure traceability and prevent the dissemination of harmful or non-consensual imagery. Furthermore, as the underlying Diffusion Transformers (DiTs) are trained on large-scale public datasets, the model may inadvertently inherit or amplify Societal Biases regarding gender, race, or cultural stereotypes. We encourage future researchers to employ de-biasing techniques and diverse datasets to foster a more equitable and responsible development of generative AI technologies.

% \begin{appendices}

% \section{Section title of first appendix}\label{secA1}

% An appendix contains supplementary information that is not an essential part of the text itself, but which may be helpful in providing a more comprehensive understanding of the research problem or it is information that is too cumbersome to be included in the body of the paper.




% \end{appendices}


\bibliography{sn-bibliography}% common bib file

\end{document}